\documentclass[letterpaper, 10 pt, conference]{IEEEtran}

\usepackage{cite}
\usepackage{graphicx}
\usepackage{subfigure}
\usepackage{algorithm}
\usepackage{algpseudocode}

\usepackage{amsmath}

\title{Night vision obstacle detection and avoidance based on Bio-Inspired Vision Sensors}

\author{\IEEEauthorblockN{Jawad N. Yasin\IEEEauthorrefmark{1},
Sherif A.S. Mohamed\IEEEauthorrefmark{1},
Mohammad-hashem Haghbayan\IEEEauthorrefmark{1},
Jukka Heikkonen\IEEEauthorrefmark{1}, 
Hannu Tenhunen\IEEEauthorrefmark{2},\\
Muhammad Mehboob Yasin\IEEEauthorrefmark{3} and Juha Plosila\IEEEauthorrefmark{1}}
\IEEEauthorblockA{\IEEEauthorrefmark{1}Department of Future Technologies, University of Turku, 20500 Turku, Finland.}
\IEEEauthorblockA{\IEEEauthorrefmark{2}Department of Industrial and Medical Electronics, Royal Institute of Technology (KTH), 16440 Kista, Sweden.}
\IEEEauthorblockA{\IEEEauthorrefmark{3}Department of Computer Networks, King Faisal University, Hofuf, Saudi Arabia.}
Emails: {\IEEEauthorrefmark{1}\{janaya, samoha, mohhag, jukhei, juplos\}@utu.fi}. \IEEEauthorrefmark{2}hannu@kth.se, \IEEEauthorrefmark{3}mmyasin@kfu.edu.sa}

\begin{document}

\maketitle
\begin{abstract}
Moving towards autonomy, unmanned vehicles rely heavily on state-of-the-art collision avoidance systems (CAS). However, the detection of obstacles especially during night-time is still a challenging task since the lighting conditions are not sufficient for traditional cameras to function properly. Therefore, we exploit the powerful attributes of event-based cameras to perform obstacle detection in low lighting conditions. Event cameras trigger events asynchronously at high output temporal rate with high dynamic range of up to 120 $dB$. The algorithm filters background activity noise and extracts objects using robust Hough transform technique. The depth of each detected object is computed by triangulating 2D features extracted utilising LC-Harris. Finally, asynchronous adaptive collision avoidance (AACA) algorithm is applied for effective avoidance. Qualitative evaluation is compared using event-camera and traditional camera.

\end{abstract}


\IEEEpeerreviewmaketitle

\vspace{-0.05cm}
\section{Introduction}

With exponential growth in the use of vehicles, the number of accidents have increased considerably, with studies showing that approximately 90\% of the accidents are due to the human error. And thereby making a reliable detection of an obstacle is one of the most important parts in advanced driver assistance systems (ADAS) or collision avoidance systems (CAS), with vision sensors among the most popular choices \cite{PRAKASH2019172, 7274498, 9108245}. Majority of the methods utilise the traditional optical sensors for detection of vehicles under normal lighting conditions such as daytime \cite{dornaikajs2016, 8764393}. Stereo vision based detection methods, motion based methods, and monocular vision detection based methods are the three kinds of methods used for obstacle detection using optical sensors \cite{5548014, lu2014vehicle}. Traditional cameras can have either indirect based methods (i.e., feature-based) or direct based methods.

\textit{Indirect \& Direct Methods:} As only some of the features can be tracked or detected, therefore the feature-based, i.e., indirect methods, are not robust when it comes to low-textured environments. However, all of the related information, even the weak intensity variations, is utilised in direct methods, making them more robust and helps in providing efficient results in similar surroundings. Since direct methods are computationally demanding, hybrid approach (which is a combination of both approaches) is used to deal with such issues. For instance, in \cite{Omni2008}, a hybrid VO approach is proposed for approximating the ground vehicle's pose. In the proposed methodology, direct method is utilised to efficiently ascertain the orientation, while feature-based technique is utilised for determining the displacement. In \cite{Feng2007}, the authors also present a direct and feature-based localization technique. In the proposed algorithm, to determine and approximate the pose, feature-based approach is used given that there are enough features in the frame. Similarly, the second part of the algorithm, i.e., direct method based module, is utilised when the environment is low-textured. Authors in \cite{Forster2014}, presented a semi-direct visual odometry approach (SVO) to tackle with the extraction of features at every frame, which is costly. In order to increase the accuracy, subpixel feature correspondence is utilised and feature extraction is applied only to the selected keyframes. However, a different hybrid approach of combining the feature-based approach with semi-dense direct image alignment is proposed by the authors in \cite{FB-DB-2}. In the proposed methodology, for keyframes, direct method is utilised, whereas indirect method is used for the other frames, and then these results are utilised for direct methods tracking.

\begin{figure*}[h]
    \centering
    \includegraphics[width=2.0\columnwidth]{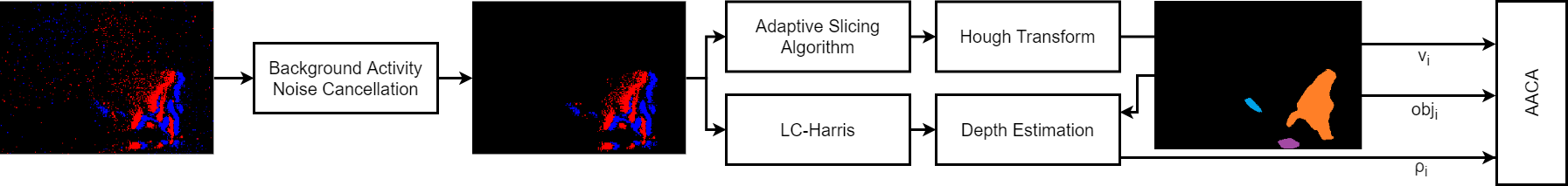}
    \caption{Overview of the overall asynchronous obstacle avoidance}
    \label{fig:sys}
    \vspace{-0.35cm}
\end{figure*}

The benefits of using event-based cameras over traditional vision sensors/cameras are: high dynamic range, low power, high temporal resolution, and low latency. Event-based cameras have significantly high dynamic range as compared to the traditional high quality frame-based cameras, i.e., 120 $dB$ vs 60 $dB$ respectively. Furthermore, in event cameras, instead of waiting for the global shutter, each pixel work independently and the photoreceptors of the pixels function in logarithmic scale. This makes event-based cameras capture information in all lighting conditions, i.e., from daytime to night time scenes.

The rest of the paper is organised as follows. Section 2 provides the development of the proposed algorithm. Results are provided in Section 3. Finally, concluding remarks and future work is given in Section 4.

\section{Proposed Methodology}

Figure \ref{fig:sys}, shows the system overview of the proposed algorithm. Which consists of four main units, i.e., noise cancellation, object detection, depth estimation, and AACA.

\vspace{-0.05cm}

\subsection{Background Activity Noise Cancellation}

Even if there is no movement or change in the brightness, background activity (BA) events are triggered because of hardware limitations, such as leakage of current in the switches or thermal noise \cite{DVS}. The generation of this noise not only deteriorates the captured data but also increases the computational costs. Therefore, in order to obtain high-performance obstacle detection and avoidance, it is crucial to have a filtering algorithm in place to eliminate the BA noise.

Our algorithm selects a service active event (SAE) size of 9x9 for each incoming event, where the central pixel is the incoming event. A kNN algorithm is used to check the correlation between the neighbouring events and the incoming events. If the incoming event does not have sufficient amount of neighbours, the filter discards the incoming event as BA noise, otherwise it is processed.

\subsection{Adaptive Slicing Algorithm \& Hough Transform}

There are various methods for slicing the incoming events for the generation of artificially synthesized events frames. In \cite{tslice}, the authors proposed a method to generate event frames by accumulating events during a fixed time interval. However, this method can generate either noisy or blurred event frames, since either highly dynamic environments or the camera motion would generate a high rate of events, which as a result generate blurry event frames. On the other hand, if the scene is static, low-rate of events are triggered and accumulating events based on time slicing would generate noisy event frames. Therefore, in our algorithm, to overcome this, we accumulate "$N$" events to generate an event frame. This number is selected based on the velocity of the objects.

The objects in the scene are detected by fitting a local plane using a randomised Hough transform \cite{ht}. Three randomised events are used to calculate the 3D of Hough space, i.e., $\theta$, $\phi$, and $\rho$. Each cell gets a vote from close events and after iterating over the points in the set, the highest voted cell is considered as the plane, i.e., the object.

\subsection{Low-complex eHarris score \& Depth Estimation}

State-of-the-art and high-performance event corner detection algorithm is eHarris. This algorithm uses Harris score \cite{harris} to detect 2D keypoints, i.e., corners, from a string of asynchronous events. However the main drawback of the mentioned algorithm is that they are computationally heavy, i.e, they demand a lot of computational power for computing the eigenvalues for all incoming events, making this method not suitable for real-time embedded systems. Furthermore, as event cameras are capable of passing on about 8 millions events per second, therefore, in order to utilise the Harris detector for systems especially with resource restrains while utilising these cameras, computational complexity reduction is crucial. Hence, we propose the more computationally friendly algorithm inspired by the eHarris, which we call, LC-Harris. 

With the size of about 9x9, we extract a binary local patch around every new occurrence of event. The most recent neighbours "\textit{N}" are considered, where \textit{N} = 25, and then given a label of 1 in the local patch. The horizontal and vertical gradients are calculated through the binary local patch, and then it is used for calculating the score as follows:

   \begin{equation}
       S = a^{'}*c^{'} = \sum|i_{x}|*\sum|i_{y}|
   \end{equation}{}
   \vspace{-0.2cm}
   
where vertical and horizontal gradients are denoted by $i_y$ and $i_x$ respectively, and $S$ represents the score. Here the incoming event is declared as a corner if the calculated score exceeds the threshold.

Using the information of the image pose, i.e., location and orientation, we can estimate the depth of 2D features, i.e., corners, using triangulation.

\subsection{Asynchronous Adaptive Collision Avoidance (AACA)}

Inspired by \cite{eof}, we estimated the relative velocity $v_i$ of the centre of mass of each object, as we consider each object as a rigid body. The distance travelled by the vehicle after $t$ seconds can be calculated as follows:

\begin{equation}
    d_v = v * (t_{i} - t_{i-1})
\end{equation}

where $d_v$ is the distance travelled by the vehicle and $v$ is the velocity of the vehicle. Then the apparent velocity of the object ($v_{obji}$) can be calculated by computing the distance travelled by the object as follows:

\begin{equation}
    d_{obji} = \rho_{i-1} - d_v - \rho_{i}
\end{equation}

\begin{equation}
    v_{obji} = d_{obji} / \Delta t
\end{equation}

where $\rho_{i-1}$ and $\rho_{i}$ are the distances between the vehicle and the object at $t_{i-1}$ and $t_{i}$ respectively. And similarly, the point of impact ($poi_i$) can be easily computed as well.

\begin{figure*}[!ht]
\begin{center}
    \subfigure[Raw events\label{fig:init1}]{\includegraphics[width=0.4\columnwidth]{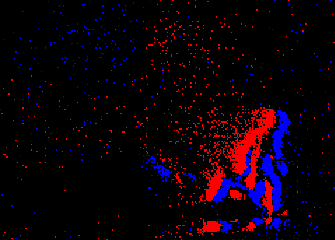}} 
    \subfigure[Filtered events\label{fig:init2}]{\includegraphics[width=0.4\columnwidth]{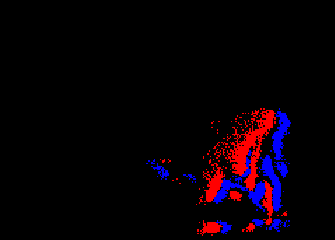}}
    \subfigure[LC\_Harris\label{fig:init3}]{\includegraphics[width=0.4\columnwidth]{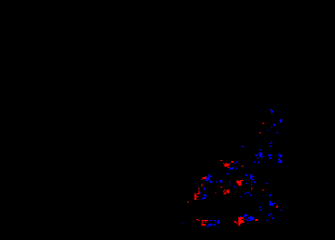}} 
    \subfigure[Object detection\label{fig:init4}]{\includegraphics[width=0.4\columnwidth]{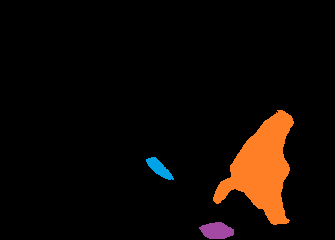}}
   \subfigure[Raw image\label{fig:init5}]{\includegraphics[width=0.4\columnwidth]{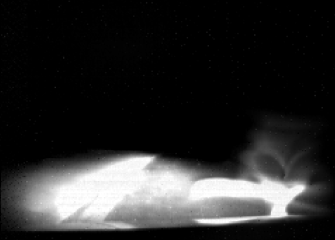}}
   \subfigure[Filtered image\label{fig:init6}]{\includegraphics[width=0.4\columnwidth]{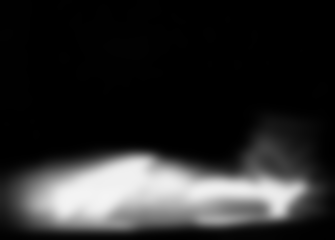}}
   \subfigure[Harris\label{fig:init7}]{\includegraphics[width=0.4\columnwidth]{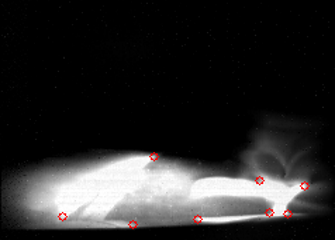}}
   \subfigure[Object detection\label{fig:init8}]{\includegraphics[width=0.4\columnwidth]{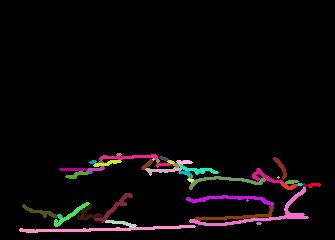}}
   \end{center}
     \caption{Qualitative comparison between event camera and traditional camera}
  \label{fig:results}
  \vspace{-0.4cm}
\end{figure*}

In the asynchronous adaptive collision avoidance module, we utilised and modified the collision avoidance algorithm proposed in \cite{10.1145/3386164.3386176} to tackle with the multi-priority obstacle switching. From the calculated values of the detected obstacles, priorities are assigned to the obstacles based on their respective $poi$, with highest priority given to the obstacle with closest $poi$ (Line 3, Algorithm \ref{algo1}). These values and the priority assignment are constantly monitored and updated (if necessary), in case a new obstacle appears in the scene or an obstacle speeds up in such a way that its $poi$ becomes the lowest. If the calculated point of impact is same for more than one object, they can be treated as being parallel objects, as at the given time of potential collision $t_i$, they both will be at the same distance from the vehicle (Line 4). Therefore, the distance or gap between the objects is calculated \cite{10.1145/3386164.3386176} and it is checked if the gap is sufficient enough to pass through, i.e., more than the defined collision radius $R_c$. In which case, the vehicle is aligned to pass through the objects safely (Lines 5-7). Otherwise, if the calculated gap between the objects is less than $R_c$, they are treated as a single obstacle and path planning is performed accordingly for successful avoidance maneuver (Lines 8-10). On the other hand, if $poi$ of the detected objects is different, then the $poi$ with highest priority is chosen for path planning and the short term path planning is done accordingly for successful avoidance maneuver (Lines 12-14).

\begin{algorithm}
\caption{Asynchronous Adaptive Collision Avoidance}\label{algo1}
\scriptsize
\begin{algorithmic}[2]
\Procedure{AACA}{$obj_i$, $v_{obji}$, $\rho_i$}
\While{obj == True}
    \State{Priority assignment($poi_i$);}
    \If{more than 1 $obj$ have same $poi$}
        \State {$D_{safe}$ $\gets$ Calculate gap between obstacles ($edges$);}
            \If{$D_{safe}$ $>$ $R_c$}
                \State{Short-term path planning ($edges$);}\Comment{Align to pass through the gap}
            \Else \Comment{Not enough gap, treat as same obstacle}
                \State{$pathPlan$ $\gets$ Calculate path plan ($edges$);}
                \State{Short-term path planning ($pathPlan$);}
            \EndIf
    \Else
        \State{$pathPlan$ $\gets$ Calculate path plan ($Priority_i$, $\rho_i$, $v_{obji}$);}
        \State{Short-term path planning ($pathPlan$);}
    \EndIf

\EndWhile
\EndProcedure
\end{algorithmic}
\end{algorithm}
\vspace{-0.4cm}

\section{Results}

Publicly available dataset of \cite{Dataset} was used for the evaluation of the proposed filtering algorithm. The selected dataset is a moving person in front of a static vehicle mounted with event camera. The dataset was recorded by a Dynamic and Active-pixel Vision Sensor DAVIS-240, which contain many sequences of frame-based, i.e., intensity images, and asynchronous events at the resolution of 240x180. Note that the intensity images are only used for comparison purposes. The proposed algorithm has been implemented in software in C++. The application was run on an Nvidia Jetson TX2 board with quad-core ARM Cortex-A57.

The qualitative comparison is summarised in Figure \ref{fig:results}, where it is evident that event camera surpass the traditional camera during night time. In event camera's output (Figure \ref{fig:init1}), a running person can be easily seen while for traditional camera's output (Figure \ref{fig:init5}) it is quite difficult to detect any movement in front. It is also evident from Figure \ref{fig:init3} and \ref{fig:init7}, that the number of extracted corners from event cameras are more accurate and robust to illumination. Three objects are extracted from the scene using our method, Figure \ref{fig:init4}, on the other hand, it is difficult to extract any objects using traditional cameras Figure \ref{fig:init8}.

\section{Conclusion}
\vspace{-0.07cm}

In this paper, we developed a night vision obstacle detection and collision avoidance algorithm utilising the dynamic vision sensor for autonomous vehicles. We performed BA filtering to eliminate noise which decreases the computational costs significantly and increases the accuracy. Then an object detection algorithm is utilised using an adaptive slicing algorithm based on accumulating number of events. Afterwards, Hough transform is used to detect objects from the generated event frames. Furthermore, the AACA (asynchronous adaptive collision avoidance) algorithm is able to detect, evaluate, and tackle with the change in environment at run-time and adapt as soon as either a new or an existing object under observation, changes its parameters, endangering the safety of the system, i.e., potential collision.

Due to the space limitation of the conference, the main emphasis of our work has been on showcasing the qualitative results. In future work, we plan to perform rigorous real-time testing under different environmental conditions to provide comprehensive qualitative and quantitative results for such DVS-based systems.

\bibliographystyle{IEEEtran.bst}
\bibliography{IEEEfull,biblio}

\begin{thebibliography}{10}
\providecommand{\url}[1]{#1}
\csname url@samestyle\endcsname
\providecommand{\newblock}{\relax}
\providecommand{\bibinfo}[2]{#2}
\providecommand{\BIBentrySTDinterwordspacing}{\spaceskip=0pt\relax}
\providecommand{\BIBentryALTinterwordstretchfactor}{4}
\providecommand{\BIBentryALTinterwordspacing}{\spaceskip=\fontdimen2\font plus
\BIBentryALTinterwordstretchfactor\fontdimen3\font minus
  \fontdimen4\font\relax}
\providecommand{\BIBforeignlanguage}[2]{{%
\expandafter\ifx\csname l@#1\endcsname\relax
\typeout{** WARNING: IEEEtran.bst: No hyphenation pattern has been}%
\typeout{** loaded for the language `#1'. Using the pattern for}%
\typeout{** the default language instead.}%
\else
\language=\csname l@#1\endcsname
\fi
#2}}
\providecommand{\BIBdecl}{\relax}
\BIBdecl

\bibitem{PRAKASH2019172}
\BIBentryALTinterwordspacing
C.~D. Prakash, F.~Akhbari, and L.~J. Karam, ``Robust obstacle detection for
  advanced driver assistance systems using distortions of inverse perspective
  mapping of a monocular camera,'' \emph{Robotics and Autonomous Systems}, vol.
  114, pp. 172 -- 186, 2019. [Online]. Available:
  \url{http://www.sciencedirect.com/science/article/pii/S0921889018301787}
\BIBentrySTDinterwordspacing

\bibitem{7274498}
P.~{Hurney}, P.~{Waldron}, F.~{Morgan}, E.~{Jones}, and M.~{Glavin}, ``Review
  of pedestrian detection techniques in automotive far-infrared video,''
  \emph{IET Intelligent Transport Systems}, vol.~9, no.~8, pp. 824--832, 2015.

\bibitem{9108245}
J.~N. {Yasin}, S.~A.~S. {Mohamed}, M.~{Haghbayan}, J.~{Heikkonen},
  H.~{Tenhunen}, and J.~{Plosila}, ``Unmanned aerial vehicles (uavs): Collision
  avoidance systems and approaches,'' \emph{IEEE Access}, vol.~8, pp.
  105\,139--105\,155, 2020.

\bibitem{dornaikajs2016}
\BIBentryALTinterwordspacing
Y.~Cai, X.~Sun, H.~Wang, L.~Chen, and H.~Jiang, ``Night-{Time} {Vehicle}
  {Detection} {Algorithm} {Based} on {Visual} {Saliency} and {Deep}
  {Learning},'' \emph{Journal of Sensors}, vol. 2016, p. 8046529, Nov. 2016,
  publisher: Hindawi Publishing Corporation. [Online]. Available:
  \url{https://doi.org/10.1155/2016/8046529}
\BIBentrySTDinterwordspacing

\bibitem{8764393}
S.~A.~S. {Mohamed}, M.~{Haghbayan}, T.~{Westerlund}, J.~{Heikkonen},
  H.~{Tenhunen}, and J.~{Plosila}, ``A survey on odometry for autonomous
  navigation systems,'' \emph{IEEE Access}, vol.~7, pp. 97\,466--97\,486, 2019.

\bibitem{5548014}
C.~{Hermes}, J.~{Einhaus}, M.~{Hahn}, C.~{Wöhler}, and F.~{Kummert}, ``Vehicle
  tracking and motion prediction in complex urban scenarios,'' in \emph{2010
  IEEE Intelligent Vehicles Symposium}, June 2010, pp. 26--33.

\bibitem{lu2014vehicle}
J.~Lu, F.~Meng-Yin, W.~Mei-Ling, and Y.~Yi, ``Vehicle detection based on vision
  and millimeter wave radar,'' \emph{Journal of Infrared and Millimeter Waves},
  vol.~33, no.~5, pp. 465--471, 2014.

\bibitem{Omni2008}
D.~Scaramuzza and R.~Siegwart, ``Appearance-guided monocular omnidirectional
  visual odometry for outdoor ground vehicles,'' \emph{IEEE Transactions on
  Robotics}, vol.~24, no.~5, pp. 1015--1026, 2008.

\bibitem{Feng2007}
J.~Feng, C.~Zhang, B.~Sun, and Y.~Song, ``A fusion algorithm of visual odometry
  based on feature-based method and direct method,'' in \emph{2017 Chinese
  Automation Congress (CAC)}, Oct 2017, pp. 1854--1859.

\bibitem{Forster2014}
C.~Forster, M.~Pizzoli, and D.~Scaramuzza, ``{SVO: Fast semi-direct monocular
  visual odometry},'' \emph{Proceedings - IEEE International Conference on
  Robotics and Automation}, pp. 15--22, 2014.

\bibitem{FB-DB-2}
N.~Krombach, D.~Droeschel, and S.~Behnke, ``{Combining feature-based and direct
  methods for semi-dense real-time stereo visual odometry},'' \emph{Advances in
  Intelligent Systems and Computing}, vol. 531, no. July, pp. 855--868, 2017.

\bibitem{DVS}
P.~{Lichtsteiner}, C.~{Posch}, and T.~{Delbruck}, ``A 128$\times$128 120 db
  15$\mu$s latency asynchronous temporal contrast vision sensor,'' \emph{IEEE
  Journal of Solid-State Circuits}, vol.~43, no.~2, pp. 566--576, Feb 2008.

\bibitem{tslice}
M.~Liu and T.~Delbruck, ``Adaptive time-slice block-matching optical flow
  algorithm for dynamic vision sensors,'' 09 2018.

\bibitem{ht}
``Randomized hough transform (rht): Basic mechanisms, algorithms, and
  computational complexities,'' \emph{CVGIP: Image Understanding}, vol.~57,
  no.~2, pp. 131 -- 154, 1993.

\bibitem{harris}
C.~Harris and M.~Stephens, ``A combined corner and edge detector,'' in
  \emph{Proceedings of the 4th Alvey Vision Conference}, 1988, pp. 147--151.

\bibitem{eof}
E.~Mueggler, C.~Forster, N.~Baumli, G.~Gallego, and D.~Scaramuzza, ``Lifetime
  estimation of events from dynamic vision sensors,'' vol. 2015, 05 2015.

\bibitem{10.1145/3386164.3386176}
\BIBentryALTinterwordspacing
J.~N. Yasin, M.-H. Haghbayan, J.~Heikkonen, H.~Tenhunen, and J.~Plosila,
  ``Formation maintenance and collision avoidance in a swarm of drones,'' in
  \emph{Proceedings of the 2019 3rd International Symposium on Computer Science
  and Intelligent Control}, ser. ISCSIC 2019.\hskip 1em plus 0.5em minus
  0.4em\relax New York, NY, USA: Association for Computing Machinery, 2019.
  [Online]. Available: \url{https://doi.org/10.1145/3386164.3386176}
\BIBentrySTDinterwordspacing

\bibitem{Dataset}
C.~Scheerlinck, N.~Barnes, and R.~Mahony, ``Continuous-time intensity
  estimation using event cameras,'' in \emph{Asian Conf. Comput. Vis. (ACCV)},
  December 2018.

\end{thebibliography}

\end{document}